\documentclass[10pt,twocolumn]{article}

\usepackage[letterpaper,top=0.67in,bottom=0.72in,left=0.67in,right=0.67in,columnsep=0.24in]{geometry}
\usepackage[T1]{fontenc}
\usepackage[utf8]{inputenc}
\usepackage{lmodern}
\usepackage{microtype}
\usepackage{graphicx}
\usepackage{balance}
\usepackage{float}
\usepackage{placeins}
\usepackage{booktabs}
\usepackage{longtable}
\usepackage{calc}
\usepackage{tabularx}
\usepackage{array}
\usepackage{ragged2e}
\usepackage{etoolbox}
\usepackage{amsmath,amssymb}
\usepackage[numbers,sort&compress]{natbib}
\usepackage[font=small,labelfont=bf,labelsep=period]{caption}
\usepackage{xcolor}
\usepackage{titlesec}
\usepackage{fancyhdr}
\usepackage[hidelinks,breaklinks=true]{hyperref}
\usepackage[nameinlink,noabbrev]{cleveref}

\definecolor{chemblue}{HTML}{26577C}
\definecolor{chemlink}{HTML}{234F70}
\hypersetup{
  unicode=true,
  pdfencoding=auto,
  colorlinks=true,
  linkcolor=chemlink,
  citecolor=chemlink,
  urlcolor=chemlink,
  pdftitle={ChemWorld: Programmable Chemical Worlds for Controlled and
Replayable Agent Experimentation},
  pdfauthor={Jiangjie Qiu; Yijun Li; Xiaonan Wang},
  pdfsubject={Programmable chemical worlds as controlled experimental
variables for agent research},
  pdfkeywords={programmable chemical worlds; world construction;
controlled counterfactual experimentation; process-complete evidence;
exact replay; autonomous chemistry}
}

\newcolumntype{L}[1]{>{\RaggedRight\arraybackslash}p{#1}}
\newcolumntype{Y}{>{\RaggedRight\arraybackslash}X}
\titleformat{\section}{\large\bfseries\color{chemblue}\raggedright\hyphenpenalty=10000\exhyphenpenalty=10000}{}{0pt}{}
\titleformat{\subsection}{\normalsize\bfseries\raggedright\hyphenpenalty=10000\exhyphenpenalty=10000}{}{0pt}{}
\titlespacing*{\section}{0pt}{10pt plus 2pt minus 1pt}{4pt}
\titlespacing*{\subsection}{0pt}{7pt plus 2pt minus 1pt}{3pt}

\providecommand{\tightlist}{%
  \setlength{\itemsep}{0pt}\setlength{\parskip}{0pt}}
\AtBeginEnvironment{thebibliography}{\small}

\title{\vspace{-1.8em}\textbf{ChemWorld: Programmable Chemical Worlds
for\\Controlled and Replayable Agent Experimentation}}
\author{%
Jiangjie Qiu\textsuperscript{1}, Yijun Li\textsuperscript{1}, Xiaonan
Wang\textsuperscript{1,*}%
\\[0.45em]
\parbox{0.92\textwidth}{\centering\small
\textsuperscript{1}Beijing Key Laboratory of Artificial Intelligence for
Advanced Chemical Engineering Materials, State Key Laboratory of
Chemical Engineering and Low-Carbon Technology, Department of Chemical
Engineering, Tsinghua University, Beijing 100084, China}%
\\[0.35em]\small *Correspondence: \texttt{wangxiaonan@tsinghua.edu.cn}%
}
\date{}

\begin{document}

\twocolumn[
\begin{@twocolumnfalse}
\maketitle
\begin{abstract}
Autonomous chemistry increasingly depends on environments in which
agents can repeatedly act, observe, and adapt. Physical laboratories
provide essential real-material evidence but are costly to repeat and
difficult to use for tightly matched interventions, whereas most digital
environments keep the underlying experimental world largely fixed. We
introduce ChemWorld, a programmable chemical environment in which
reusable process and observation components are compiled into executable
worlds. ChemWorld separates the public experimental contract available
to an agent from evaluator-owned chemical and material laws. Researchers
can therefore vary world composition and operating conditions, or change
a single hidden law while holding the public task and interaction
conditions fixed. Transactional execution records operations, failures,
resource changes, and state transitions, allowing complete
environment--action trajectories to be replayed exactly and audited.
Full-census qualification covered the reference registry, 52 generated
compositions, and module, interface, compilation, and invalid-action
tests. Eight deterministic experimental cases demonstrated shared
lifecycle semantics, failure recovery, and exact replay, while six
parent--child world-fork pairs isolated the effects of single
private-law interventions under matched public conditions. An
independent agent also completed a full lifecycle in a non-reference
world through the same public interface. Within the declared component
and model domain, ChemWorld provides a controlled and replayable
substrate for studying experimentation across systematically varied
chemical worlds, complementary to physical-laboratory evidence and
calibration.
\end{abstract}
\vspace{0.6em}
\end{@twocolumnfalse}
]

\section{1. Introduction}\label{introduction}

Recent advances in self-driving laboratories and chemistry agents are
moving chemical research from algorithmic assistance toward autonomous
experimentation. Agents can now plan experiments, call chemistry tools
and interact with robots, automated instruments and cloud laboratories
to execute synthesis, characterization and optimization workflows on
real materials
\citep{boiko2023autonomous, bran2024augmenting, szymanski2023alab, dai2024mobile, darvish2025organa, song2025chemagents, panapitiya2026autolabs, pilon2026robochemflex, vriza2026instruments}.
As agents assume a sustained role in experimental decisions, their
capabilities depend not only on the model but also on the environment
that defines available operations, observations and state transitions.
Building environments for agent interaction, controlled study and
evaluation is therefore becoming a foundational problem in autonomous
chemistry.

Physical laboratories provide the most direct evidence from real
materials and instruments, but large-scale repeated interaction is
costly: it consumes reagents, samples and instrument time and is
constrained by equipment, safety requirements and experimental
turnaround. At the same time, physical experiments are difficult to
restart from exactly the same initial state or to support strict
comparisons in which only one factor changes and all others remain
fixed. Software environments offer lower-cost repeatable interaction,
and existing systems span reaction optimization and process control,
virtual chemistry, interactive scientific discovery and embodied
laboratory simulation
\citep{felton2021summit, hase2021olympus, bloor2024pcgym, beeler2024chemgymrl, jansen2024discoveryworld, gandhi2025boxinggym, duan2025scigym, nagele2026sciexplorer, zheng2026newtonbench, malik2026made, li2025labutopia, wu2026labimus, xu2026scidisco}.

However, these environments usually define the experimental world in
advance and primarily study how an agent completes a given task,
operates a particular bench or adapts to fixed dynamics and hidden
rules. It is less common to systematically alter the chemical processes
and underlying laws that constitute the environment while keeping the
agent-facing operations, instruments and observations unchanged. This
limits another important class of controlled studies: under identical
public experimental conditions, one may wish to change only a material
property or process law and observe the resulting changes in the
experimental process and agent behavior. Supporting such studies
requires not just more predefined tasks, but an environment in which
chemical worlds themselves can be constructed, modified, repeated and
compared.

\begin{figure*}[!t]
\centering
\includegraphics[width=\textwidth]{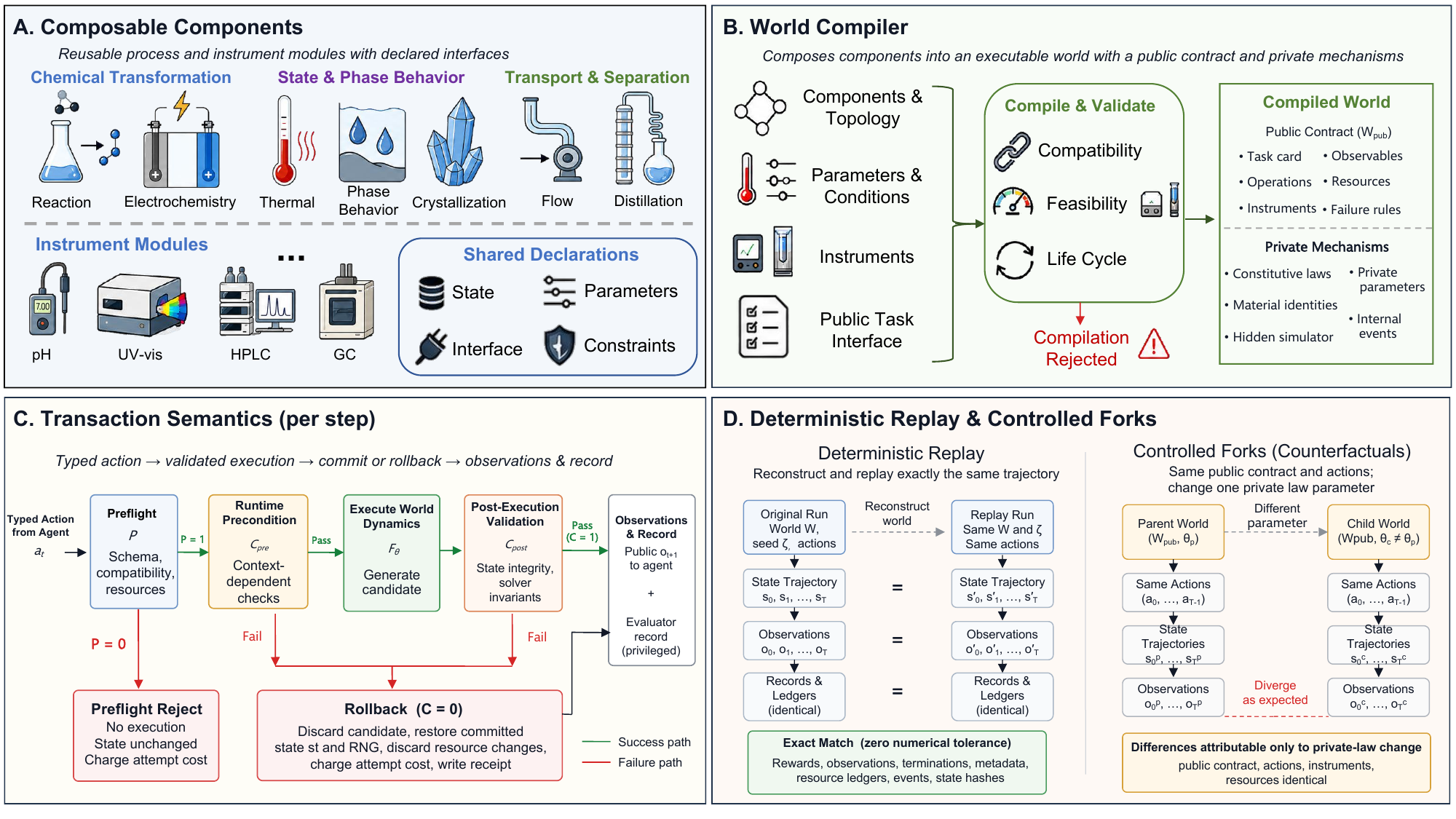}
\caption{\textbf{ChemWorld composes declared modules into executable worlds with transactional semantics, deterministic replay and controlled forks.}
\textbf{A,} Reusable process and instrument modules expose shared declarations.
\textbf{B,} The compiler either produces a public contract $W_{\mathrm{pub}}$ with evaluator-owned private mechanisms or rejects an invalid composition before construction.
\textbf{C,} Typed actions traverse preflight, runtime-precondition, candidate-execution and post-execution validation gates; non-commit branches preserve committed state and record declared attempt consequences.
\textbf{D,} Replay reconstructs the bound world and submitted action trace, while controlled forks hold the public contract and actions fixed and change one private law.}
\label{fig:overview}
\end{figure*}

We therefore introduce ChemWorld, a programmable chemical-world platform
for autonomous chemistry research (Fig. \ref{fig:overview}). ChemWorld
treats world construction as a locus of experimental control: it
represents chemical processes and observation models as reusable
components and separates the agent-accessible public experimental
interface from evaluator-owned chemical and material laws. Researchers
can compose processes, change operating conditions and instrument
configurations, or alter one hidden law while holding the public task,
action sequence and bound randomness fixed. The system records
operations, state transitions, resource use, failures and recovery,
thereby supporting repeatable, auditable and attributable experiments.

This paper makes four contributions:

\begin{enumerate}
\def\labelenumi{\arabic{enumi}.}
\tightlist
\item
  \textbf{Programmable chemical-world construction.} Reusable process
  and observation components and a compatibility compiler construct
  executable worlds whose topology, operating conditions, instruments
  and private chemical or material laws can be varied systematically.
\item
  \textbf{A common experimental interface across worlds.} Compatible
  worlds share one agent-facing contract for operations, instruments,
  observations, resources, failure handling, termination and evaluation.
\item
  \textbf{Complete, replayable and attributable experimental records.}
  Each operation executes transactionally and records success, failure,
  rollback and resource changes, providing process-complete evidence for
  replay and intervention attribution. Complete environment--action
  traces can be reconstructed exactly, while single-private-law forks
  isolate the effect of a registered world change.
\item
  \textbf{Agent experimentation through the same environment.}
  Deterministic workflows and an independent agent use the same public
  interface while the evaluator retains the complete experimental
  process record without exposing private state.
\end{enumerate}

This work establishes ChemWorld's construction, qualification and
agent-interface integration within the declared component and
compatibility domain; comparative agent performance, universal chemical
fidelity and physical-laboratory transfer remain outside the present
scope.

\textbf{Open-source release.} Executable code, frozen protocols,
processed evidence, replayable trajectories, figure sources and the
manuscript release are openly available in the
\href{https://github.com/sunyrain/ChemWorld-Public}{ChemWorld public
repository}. The frozen code-and-evidence snapshot referenced in this
paper is versioned as
\href{https://github.com/sunyrain/ChemWorld-Public/tree/v0.1.0}{v0.1.0}.

\section{2. Relation to existing experimental
environments}\label{relation-to-existing-experimental-environments}

Experimental environments for autonomous science have developed along
four related lines with different primary emphases: autonomous
experimentation in physical laboratories, laboratory simulation for
embodied agents, interactive environments for scientific discovery, and
virtual chemistry and process environments. We discuss these directions
in turn to position ChemWorld within the existing landscape.

\subsection{2.1 Autonomous experimentation in physical
laboratories}\label{autonomous-experimentation-in-physical-laboratories}

Coscientist and ChemCrow connect language-model reasoning with chemistry
tools, automated platforms and cloud laboratories. A-Lab, mobile robotic
systems, ORGANA, ChemAgents and newer instrument-facing agents further
demonstrate closed-loop synthesis, characterization, long-horizon
experimental workflows and autonomous operation of scientific
instruments on real materials
\citep{boiko2023autonomous, bran2024augmenting, szymanski2023alab, dai2024mobile, darvish2025organa, song2025chemagents, panapitiya2026autolabs, pilon2026robochemflex, vriza2026instruments}.
These systems establish the real-material execution branch of autonomous
chemistry. Their strength is direct physical experimental evidence,
although large-scale, strictly matched repetition remains limited by
samples, instrument access and experimental turnaround.

\subsection{2.2 Simulated laboratories for embodied scientific
agents}\label{simulated-laboratories-for-embodied-scientific-agents}

LabUtopia and Labimus use laboratory simulation to train and evaluate
embodied scientific agents, covering scene understanding, object and
material manipulation, instrument use and long-horizon experimental
procedures \citep{li2025labutopia, wu2026labimus}. These environments
primarily study how an agent carries out experimental procedures,
reducing the cost and risk of training directly on physical hardware and
complementing environments centered on scientific reasoning or
chemical-process dynamics.

\subsection{2.3 Interactive environments for scientific discovery
agents}\label{interactive-environments-for-scientific-discovery-agents}

DiscoveryWorld, BoxingGym, SciGym, SciExplorer and NewtonBench organize
scientific problems as sequential cycles of hypothesis formation,
experiment selection, observation and inference. They evaluate whether
agents can acquire evidence actively and recover scientific rules in
initially unknown systems
\citep{jansen2024discoveryworld, gandhi2025boxinggym, duan2025scigym, nagele2026sciexplorer, zheng2026newtonbench}.
MADE extends this paradigm to budget-constrained closed-loop materials
discovery \citep{malik2026made}, while SciDisco uses process-verifiable
scientific-discovery environments for agentic reinforcement learning and
assigns turn-level training signals to intermediate actions that produce
verifiable evidence \citep{xu2026scidisco}. Together, these works
establish interactive scientific environments as infrastructure for both
agent evaluation and training.

\subsection{2.4 Virtual chemistry and process
environments}\label{virtual-chemistry-and-process-environments}

The closest prior work to ChemWorld lies in virtual chemistry and
process environments. Summit and Olympus provide repeatable software
benchmarks for reaction optimization and experiment planning, while
PC-Gym extends this approach to chemical-process control with nonlinear
dynamics, constraints and disturbances
\citep{felton2021summit, hase2021olympus, bloor2024pcgym}. ChemGymRL
constructs a modular virtual chemistry laboratory in which reaction,
extraction, distillation and characterization are organized as
interactive chemical benches. Stable action and observation interfaces
support reinforcement-learning training and policy comparison
\citep{beeler2024chemgymrl}.

ChemGymRL and ChemWorld both expose chemical processes through
interactive software environments, but they operate at different levels
of control. ChemGymRL makes virtual chemical benches configurable for
agent training and evaluation; ChemWorld instead treats the composition
and hidden laws of the chemical world itself as controlled experimental
variables while preserving a common public agent interface. Researchers
can therefore alter the underlying world while holding the task and
interaction conditions fixed for strictly matched comparison.
Transactional execution and evaluator-complete records further make
these matched world changes exactly replayable and experimentally
attributable. Table \ref{tab:related-position} summarizes this
functional rather than performance-based distinction in terms of
repetition, experimental intervention and record scope.

\begin{table*}[!t]
\centering
\small
\caption{\textbf{Complementary operating regimes for autonomous chemical experimentation.} ChemWorld combines software-scale repetition with explicit world composition, private-law intervention and process-complete replay.}
\label{tab:related-position}
\begin{tabularx}{\textwidth}{@{}L{0.18\textwidth}L{0.18\textwidth}L{0.20\textwidth}L{0.20\textwidth}Y@{}}
\toprule
System class & Representative emphasis & Replication regime & Experimental intervention & Observable record \\
\midrule
Physical SDL / chemistry robot & real-material execution and hardware integration & apparatus-, material- and time-bound physical repeats & protocol, material and hardware changes & sensor, automation and sample records \\
Embodied laboratory simulator & procedural and manipulative competence & resettable simulated episodes & scenes, objects, procedures or embodiment configurations & actions, observations and simulator state \\
Scientific-discovery environment & evidence gathering, hypothesis testing and law recovery & resettable interactive tasks & datasets, hypotheses, latent rules or oracle budgets & analyses, actions, observations and evidence \\
Optimization / virtual chemistry environment & objective optimization, process control and interactive chemistry & software-scale queries or control episodes & objectives, process models, bench configurations or operating conditions & objective histories, controller traces or task state \\
ChemWorld & controlled experiments over composable executable worlds & exact reset, matched repetition and version-bound replay without direct wet-laboratory consumable use & declared component composition plus single-private-law forks under an invariant public contract & typed actions, evaluator-complete state, public observations, failures, resources, termination, lineage and exact environment replay \\
\bottomrule
\end{tabularx}
\end{table*}

\section{3. Programmable chemical world
construction}\label{programmable-chemical-world-construction}

ChemWorld is designed not as another collection of fixed tasks, but as
an experimental medium in which chemical worlds can be constructed and
modified systematically. It first represents chemical processes as
reusable components, compiles compatible declarations into executable
worlds, and then executes every operation through one transactional
runtime. This section defines that construction stack: the world
representation, the compatibility compiler and the execution and replay
semantics shared by all compiled worlds.

\subsection{3.1 Composable chemical
worlds}\label{composable-chemical-worlds}

ChemWorld represents reaction, thermal, phase, separation,
crystallization, distillation, continuous-flow, electrochemical and
observation processes as reusable components. Each component declares
its parameter domains, dependencies, owned state and public interfaces.
Researchers define a chemical world by selecting components, specifying
their topology and binding operating conditions and instruments, rather
than implementing a separate environment for every task. Reaction and
electrochemical components transform material state; thermal, phase,
separation, crystallization, distillation and continuous-flow components
supply process-specific transitions; and observation components attach
synthetic instruments. All public operations are typed actions rather
than unrestricted simulator mutations.

A complete world is represented as

\[
\mathcal{W}=(W_{\mathrm{pub}},\theta),\qquad
T=(W_{\mathrm{pub}},S_{0,\mathrm{pub}},A,I,O,R,\tau,E).
\]

Here \(W_{\mathrm{pub}}\) contains the public component topology, public
parameter domains and interfaces, whereas \(\theta\) contains
evaluator-owned constitutive laws, material properties, hidden
parameters and private initialization. The public task contract \(T\)
joins the public world description and initial-state projection with the
allowed actions, instruments, observations, resources, termination rule
and evaluation rule. The complete world identity therefore binds both
\(W_{\mathrm{pub}}\) and \(\theta\), while the agent interacts only
through \(T\).

We distinguish three identity levels so that task identity, world
identity and randomness do not silently serve interchangeable roles:

\[
\begin{aligned}
\mathrm{world\text{-}spec\ ID}
  &=\operatorname{id}(W_{\mathrm{pub}},\theta),\\
\mathrm{scenario\ ID}
  &=\operatorname{id}(W_{\mathrm{pub}},\theta,\zeta_{\mathrm{init}},
    \zeta_{\mathrm{dyn}},\zeta_{\mathrm{obs}}),\\
\mathrm{task\text{--}world\ unit}
  &=(\mathrm{task\ ID},\mathrm{scenario\ ID}).
\end{aligned}
\]

The qualification study uses the final level when testing overlap with
the reference registry: a generated row overlaps only if both its task
and scenario identities match a registered unit. This public/private
separation also makes a second operation possible. A researcher can
preserve the agent-facing task contract while changing one registered
element of \(\theta\), turning a private world law into a controlled
experimental variable.

\subsection{3.2 Compatibility compilation and experimental
contracts}\label{compatibility-compilation-and-experimental-contracts}

A world declaration specifies component selection and topology, public
parameters, instrument configuration and the agent-facing task surface.
ChemWorld does not execute this recipe directly. Its compatibility
compiler first normalizes the declaration and then checks component
dependencies, state ownership, unit agreement, supported parameter
domains, resource feasibility, instrument availability, operation
exposure and lifecycle closure.

Only declarations that satisfy all checks are compiled into executable
worlds. An invalid declaration fails closed before environment
construction and returns structured diagnostics, rather than leaving a
partially constructed or semantically ambiguous environment. A
successful compilation returns both the executable world and one public
experimental contract that exposes the allowed typed operations,
instruments, observations, resources, failure semantics, termination and
evaluation, while evaluator-owned mechanisms and hidden state remain
private.

Composability therefore does not mean arbitrary juxtaposition. A new
world enters the runtime only when its component interfaces, resource
conditions and lifecycle jointly satisfy the declared compatibility
rules. Conversely, an additional process model or an empirically
calibrated formulation can enter through the same interfaces without
redesigning the public task contract, transaction layer or replay
machinery. Compilation thus separates world authoring from both
process-model implementation and agent integration.

\subsection{3.3 Transactional execution and exact
replay}\label{transactional-execution-and-exact-replay}

Once a world has been compiled, every submitted action follows the same
transactional sequence: preflight admission, runtime-precondition
evaluation, candidate execution, post-execution validation and either
commit or rollback. A schema, compatibility and resource predicate
\(P(s_t,a_t,R_t)\) first determines whether an action may enter the
runtime. An admitted action is then tested against its context-dependent
runtime preconditions. If those preconditions pass, the bound mechanism
\(\theta\), committed state \(s_t\), resource ledger \(R_t\), action
\(a_t\) and recorded random variates \(\xi_t\) generate a candidate
transition:

\[
(\tilde{s}_{t+1},\tilde{R}_{t+1},\tilde{e}_{t+1})=
F_\theta(s_t,R_t,a_t,\xi_t).
\]

Candidate state is not installed immediately. A runtime commit-gate
predicate \(C\in\{0,1\}\) covers both a runtime-precondition rejection
before candidate generation and post-execution checks of state
integrity, solver status, runtime invariants and the observation path.
For a generated candidate, the latter gate is written
\(C(\tilde{s}_{t+1},\tilde{R}_{t+1},\tilde{e}_{t+1})\). The transaction
commits only when admission and all applicable runtime checks pass:

\[
\begin{aligned}
P=1,\ C=1 &\quad\Longrightarrow\\[-0.15em]
(s_{t+1},R_{t+1},e^\star_{t+1})
&=(\tilde{s}_{t+1},\tilde{R}_{t+1},e^{\mathrm{acc}}_{t+1}).
\end{aligned}
\]

The two non-commit branches remain distinct in the record. If \(P=0\),
the runtime emits a preflight-rejection event and receipt without
runtime execution. If \(P=1\) but \(C=0\), it emits a runtime-rollback
event and receipt; this branch can occur before candidate generation at
a declared runtime precondition or after candidate generation at a
post-execution check. For \(b\in\{\mathrm{pre},\mathrm{roll}\}\),

\[
s_{t+1}=s_t,\qquad
R_{t+1}=G_b(R_t,a_t,e^b_{t+1}),\qquad
e^\star_{t+1}=e^b_{t+1}.
\]

The branch-specific ledger function installs only the protocol-declared
attempt cost or penalty. Candidate physical, observation and uncommitted
resource effects are discarded. The committed runtime state also binds
the observation-RNG state \(\rho_t\); a non-commit branch restores
\(\rho_t\), preventing an unsuccessful attempt from changing future
observation noise. Public and evaluator records are separate projections
of the realized branch,

\[
o_{t+1}=\pi_{\mathrm{pub}}(s_{t+1},e^\star_{t+1}),\qquad
\hat{o}_{t+1}=\pi_{\mathrm{eval}}(s_{t+1},e^\star_{t+1}).
\]

Exact replay binds the normalized contract, runtime, mechanism and
scoring identities, together with the seeds and intervention record. It
reconstructs the compiled world and resubmits the full submitted
action/transaction trace---including committed actions, preflight
rejections and runtime rollbacks---and compares public observations,
transaction outcomes, affected-ledger declarations, world events,
rewards and terminal flags at zero numerical tolerance. Resource deltas
and rollback receipts are reconciled separately. Replay therefore
reconstructs a version-bound environment/action trajectory, not merely
its endpoint or action list, and is distinct from experience replay used
to train a reinforcement-learning policy.

\section{4. Qualification of composed chemical
worlds}\label{qualification-of-composed-chemical-worlds}

Programmability is useful only if newly composed worlds preserve the
same executable contract. We therefore qualified ChemWorld at four
connected levels: construction coverage, complete-world execution,
module and interface semantics, and failure, resource and replay
behavior. The design tests a finite declared construction domain; it
does not claim exhaustive coverage of chemical space or all higher-order
process interactions.

\subsection{4.1 Qualification design and construction
coverage}\label{qualification-design-and-construction-coverage}

The public capability map contains 15 registered reference tasks, 28
typed operation kinds and five synthetic instrument contracts. These
tasks anchor interpretable examples but do not define the boundary of
the executable world space. The reference qualification set contains 64
task--world units and 1,786 boundary and categorical recipes. To test
construction beyond these reference examples, an execution protocol
fixed eight component patterns and generated 52 additional compositions
before authoritative qualification.

The generated set separates two forms of expansion. Eighteen
compositions use three topologies absent from the reference registry:
phase--observation, phase--separation--observation and
reaction--thermal--continuous-flow--observation. Eight
reaction--thermal--distillation--observation compositions reuse a
registered topology but have zero exact task--world identity overlap
with the frozen registry. These are protocol-frozen non-reference
compositions. The remaining 26 rows provide additional coverage within
registered topologies. The first row of the eight-case non-reference
block was fixed in advance as the agent-integration target, rather than
selected after observing its outcome.

Coverage selection combined pairwise rows for discrete component and
instrument interactions, seeded Latin hypercube sampling for authored
continuous domains, and ordered workflow targets for critical process
sequences. The generated suite attained all registered targets: 60/60
discrete levels, 180/180 compatible discrete pairs, 212/212 continuous
strata and 84/84 ordered workflow interactions. These denominators
define a finite qualification sample within declared domains. They are
not evidence of semantic completeness over all higher-order chemistry;
higher-order behavior is tested only where it appears explicitly in a
workflow, module probe or interface path.

\begin{figure*}[!t]
\centering
\includegraphics[width=\textwidth]{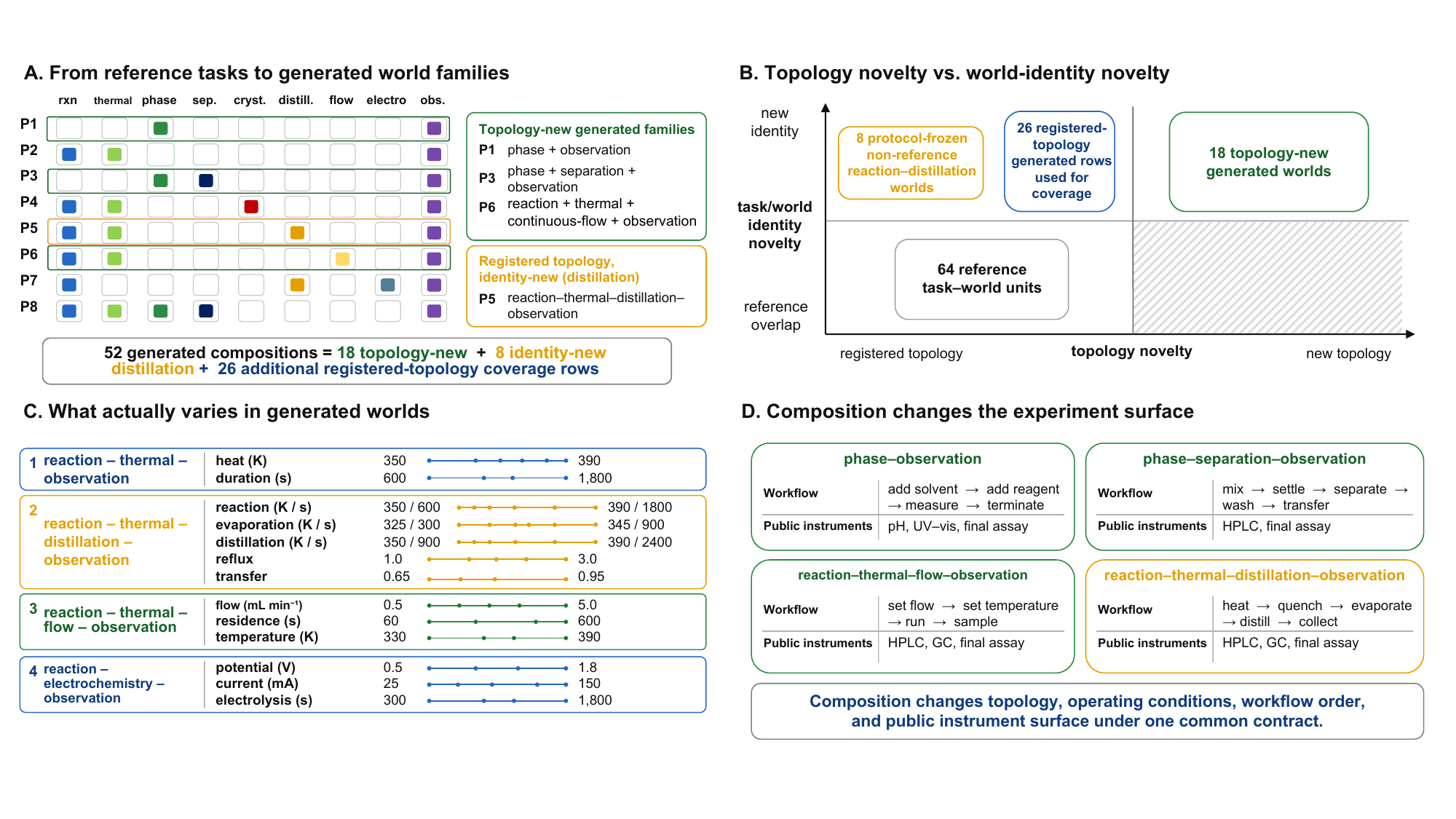}
\caption{\textbf{Construction coverage and qualification design.}
\textbf{A,} The 52 generated compositions separate into 18 topology-new worlds, eight identity-new reaction--distillation worlds that reuse a registered topology, and 26 additional registered-topology coverage rows.
\textbf{B,} Topology novelty and exact task--world identity novelty are independent coordinates relative to the 64 reference task--world units.
\textbf{C,} Protocol-frozen rows vary discrete component patterns, authored continuous operating conditions and ordered workflows.
\textbf{D,} Composition changes topology, operating conditions, workflow ordering and the public instrument surface within one common contract structure.
Coverage defines a finite construction sample rather than the extent of the world-design space; qualification outcomes are reported in Sections 4.2--4.3 and Appendix C.}
\label{fig:qualification}
\end{figure*}

\subsection{4.2 World, module and interface
qualification}\label{world-module-and-interface-qualification}

Each task--world unit had to compile, execute a complete workflow, close
its lifecycle exactly once, reconcile declared and observed resources,
and replay exactly. All 64/64 reference units passed, and the boundary
and categorical set produced 1,786/1,786 complete executions. All 52/52
generated compositions also passed, including all 8/8 protocol-frozen
non-reference reaction--distillation worlds. There were no missing
receipts, failure classes or undeclared private-field-exposure findings.
Compilation alone was never a pass condition.

Qualification also tested the construction boundary. Seven deliberately
invalid declarations covered missing dependencies, conflicting state
ownership, unit mismatches, invalid parameter domains, resource
impossibility and lifecycle gaps. All 7/7 failed closed before
environment construction with the registered diagnostic class. The
valid-world census and invalid-world mutants therefore test both sides
of the compiler boundary.

Complete-world success depends in turn on the meaning of component
transitions. Thirty-two module probes exercise zero input, declared
parameter boundaries, monotonic directions, conservation and
model-specific invariants. Seven cross-module paths test whether
material amount, unit, identity and state meaning survive transfer
between components and, where applicable, also test charge, energy and
phase balance and event propagation. All 32/32 module probes and 7/7
interface paths passed.

These results establish internally coherent executable formulations
within their declared model-card domains. Their numerical fixtures and
directional oracles are internal qualification checks, not independent
reference implementations or external physical validation. The
interfaces identify where alternative or empirically calibrated models
may be introduced while retaining the public task contract, transaction
layer and replay machinery.

\subsection{4.3 Failure semantics, resource accounting and replay
qualification}\label{failure-semantics-resource-accounting-and-replay-qualification}

The mechanisms defined in Section 3.3 must also hold under invalid
actions and exhausted resources. The frozen campaign contained 192
negative probes: 64 invalid-schema/unknown-operation probes, 64
campaign-resource-exhaustion probes and 64 runtime-precondition probes.
Every probe produced its registered outcome and preserved committed
physical state. The evidence therefore qualifies 128 \(P=0\) admission
rejections and 64 \(P=1,C=0\) runtime-precondition rollbacks.
Solver-diagnostic and candidate-observation fault paths are implemented
as fail-closed semantics, but this campaign did not assign them separate
qualification denominators.

\begin{table}[!b]
\centering
\scriptsize
\caption{\textbf{Reader-visible branch census for the 192 negative probes.} Counts are inherited from the frozen qualification report; no post hoc probe reclassification was used.}
\label{tab:transaction-branch-census}
\begin{tabularx}{\columnwidth}{@{}L{0.16\columnwidth}rY@{}}
\toprule
Branch & $n$ & Registered class and outcome \\
\midrule
$P=0$ & 64 & schema / unknown operation; validation rejection before runtime execution \\
$P=0$ & 64 & campaign-resource exhaustion; rejection before candidate installation \\
$P=1,\ C=0$ & 64 & runtime precondition failure; rollback preserves committed physical and observation-RNG state \\
\bottomrule
\end{tabularx}
\end{table}

The resource ledger independently tracks material, sample, instrument
use, process time, operation count and terminal assay. Failed attempts
retain only their declared costs or penalties, and observation checks
require public packets to contain only task-declared fields. Exact
replay then reconstructs the bound compiled world and resubmits the
complete typed-action sequence, including accepted actions, admission
rejections and runtime rollbacks. It compares rewards, public
observations, termination flags, transaction metadata, affected-ledger
declarations, world events, state-delta summaries and state-integrity
checks with zero numerical tolerance; case qualification separately
reconciles ledger deltas and rollback receipts. All reported replays
yielded zero numerical mismatch. Thus the qualified reproducible unit is
the complete environment/action trajectory, including failure and
resource consequences, rather than an endpoint or simplified action
list.

\section{5. Controlled experiments over chemical
worlds}\label{controlled-experiments-over-chemical-worlds}

Qualification establishes that ChemWorld constructs and executes worlds
consistently. We next ask what controlled experiments this medium
supports. The first study tests process-complete lifecycles, including
planned failure and recovery, across diverse workflows. The second makes
world construction itself experimental by changing one private law under
otherwise matched public conditions.

\subsection{5.1 Process-complete lifecycles and failure
recovery}\label{process-complete-lifecycles-and-failure-recovery}

Eight protocol-frozen use cases span reaction-to-crystallization,
resource-limited equilibrium characterization, planned failure and
recovery, continuous flow, electrochemistry, distillation, partition and
a second crystallization world. Together they cover single-stage and
multistage processing, constrained measurement and multiple separation
modalities. Each case is an independent experimental unit whose
submitted actions are audited within one complete lifecycle.

Across the eight cases, all 89 submitted actions have complete schema,
transaction, state-integrity, event, resource and public-observation
receipts. Eighty-eight actions committed and one protocol-frozen action
rolled back. Every case completed one final assay, closed its lifecycle,
reconciled resources and replayed exactly with zero numerical error. The
same lifecycle contract therefore supports distinct chemical workflows
without process-specific failure, termination or recording semantics.

The failure--recovery case places one deliberate invalid operation
inside an otherwise complete experiment. Its first action passed schema,
compatibility and campaign-resource admission, but a
runtime-precondition check found that no separable phase had yet formed.
The transaction consequently entered the recorded \(P=1,C=0\) branch
before candidate physical state was generated. It preserved committed
physical state and observation-RNG state, created no ghost state, and
reconciled the declared failed-attempt consequences. The next 18 actions
continued from the last committed state and completed the recovery path,
final assay, resource reconciliation and exact replay of all 19
submitted actions.

Failure is therefore an experimental event rather than an episode-level
exception. Its attempt and resource consequences remain auditable, while
rejected physical state is excluded from subsequent execution. A
workflow can recover without erasing committed history or restarting the
experimental unit.

\subsection{5.2 Controlled single-law
counterfactuals}\label{controlled-single-law-counterfactuals}

Process-complete execution and controlled attribution require different
designs. The lifecycle study asks whether diverse workflows share one
execution semantics; the counterfactual study asks whether one private
world law can change while all public experimental conditions remain
fixed. Parent and child worlds are defined as

\[
\begin{aligned}
\mathcal{W}_{p}&=(W_{\mathrm{pub}},\theta_p),&
\mathcal{W}_{c}&=(W_{\mathrm{pub}},\theta_c),\\
\theta_p&\neq\theta_c,& T_p&=T_c=T.
\end{aligned}
\]

The qualification contains six parent--child pairs: two intervention
classes evaluated over three seeds. Every pair preserves nine versioned
public-contract components---task, actions, instruments, observations,
resources, failures, scoring, material catalogue and the contracted
invariant/safety surface---as well as the fixed typed-action sequence
and bound randomness. Parent and child consequently have distinct
complete-world identities while presenting the same public experiment.
The admissible change is restricted to exactly one protocol-frozen
private constitutive or material law.

The partition intervention changes the hidden response from \(K^{1.00}\)
to \(K^{1.75}\); the registered terminal organic-product amount and
public \texttt{product\_in\_organic} assay both increase. The
electrochemical intervention keeps public material labels fixed while
reassigning hidden electrolyte-response profiles; the registered
selective-product amount and public \texttt{ohmic\_efficiency} both
decrease. For each intervention, direction is checked separately by the
frozen divergence oracle, while the magnitude oracle requires both the
protocol-frozen absolute and relative thresholds.

Repeating both variants produced 24 deterministic traces. All six pairs
passed lineage, exactly-one-private-target, public-contract-invariance,
same-sequence-executability, expected-state-divergence,
expected-observation-divergence and exact-replay gates. Within the
declared executable model and intervention domain, the resulting
trajectory differences are therefore attributable to the registered
private-law change under fixed actions and noise identity.

\begin{table*}[!t]
\centering
\scriptsize
\caption{\textbf{Reader-visible specification of the two controlled-fork classes.} Public task, actions, instruments, observations, resources, failures, scoring, material catalogue and the contracted invariant/safety surface remain unchanged in every row. Thresholds were frozen before execution and require both the listed absolute and relative magnitudes plus the stated direction.}
\label{tab:fork-specification}
\begin{tabularx}{\textwidth}{@{}L{0.12\textwidth}L{0.15\textwidth}L{0.25\textwidth}L{0.19\textwidth}Y@{}}
\toprule
Fork class & Private target & Parent $\rightarrow$ child law & Registered state channel & Registered public channel \\
\midrule
Partition constitutive law & phase-partition response & partition-base response $K^{1.00}\rightarrow K^{1.75}$ & terminal organic-product amount $P_{\mathrm{org}}$ increases; $\Delta\geq10^{-4}$ mol and relative difference $\geq0.05$ & final-assay \texttt{product\_in\_organic} increases; $\Delta\geq0.02$ and relative difference $\geq0.02$ \\
Electrochemical material law & hidden electrolyte-profile effects & public profile labels stay fixed; hidden effect-row mapping $(0,1,2,3)\rightarrow(2,1,0,3)$ & terminal selective-product amount \texttt{Red} decreases; $\Delta\geq10^{-6}$ mol and relative difference $\geq0.01$ & final-assay \texttt{ohmic\_efficiency} decreases; $\Delta\geq0.05$ and relative difference $\geq0.05$ \\
\bottomrule
\end{tabularx}
\end{table*}

In the electrochemical fork, the permutation reassigns hidden response
profiles to unchanged public material labels. It is therefore a matched
change in private material properties, not an identifier remapping or a
change in the public catalogue.

\begin{figure*}[!t]
\centering
\includegraphics[width=\textwidth]{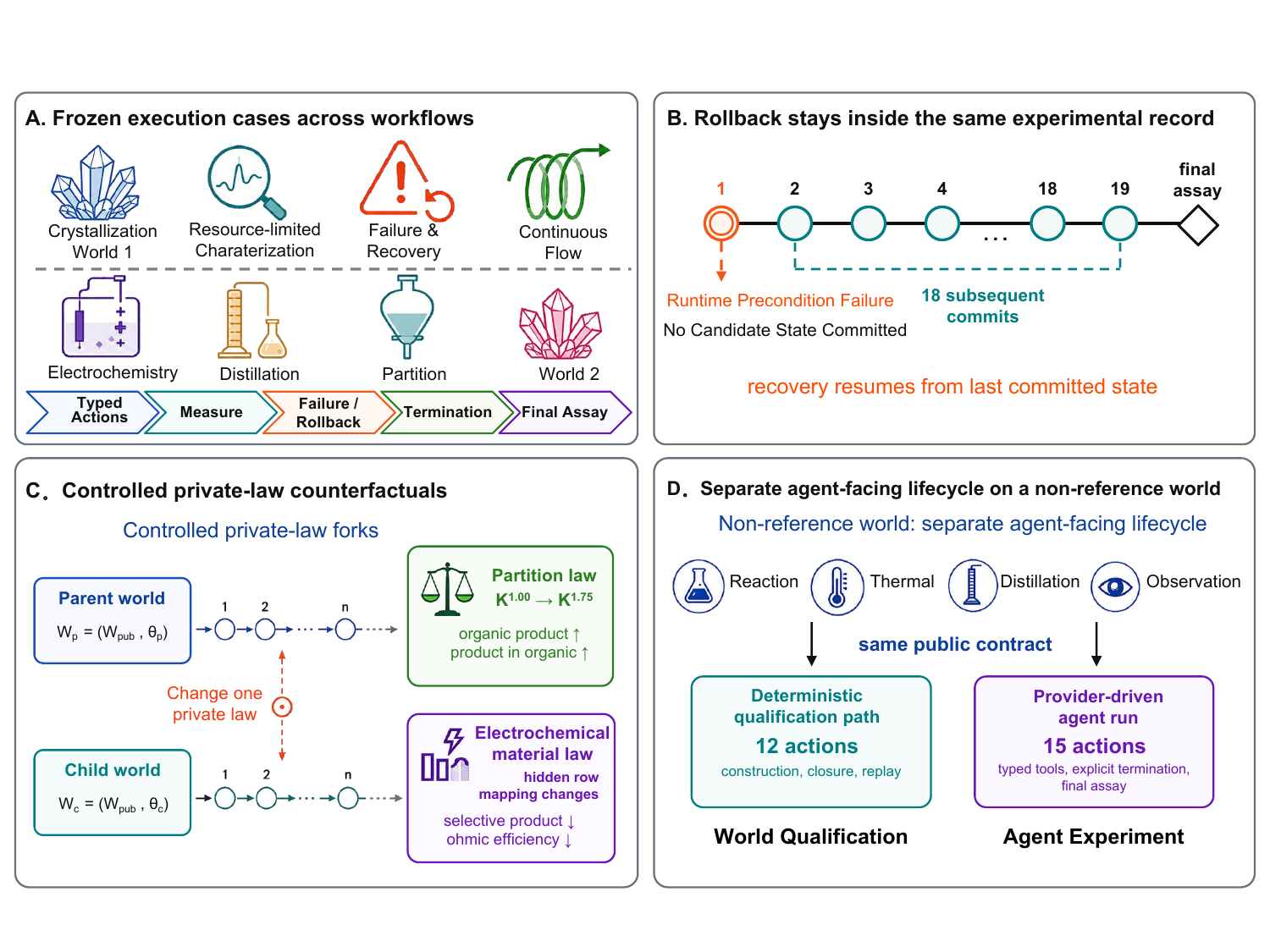}
\caption{\textbf{Execution, intervention and agent access.}
\textbf{A,} Eight frozen execution cases span crystallization, resource-limited characterization, planned failure and recovery, continuous flow, electrochemistry, distillation and partition workflows under one shared lifecycle semantics.
\textbf{B,} A runtime-precondition failure at step 1 remains inside the same experimental record; recovery continues from committed state through 18 subsequent commits and final assay.
\textbf{C,} Controlled private-law forks hold the public contract, typed action sequence and bound randomness fixed while changing one private law.
\textbf{D,} A separate non-reference world supports deterministic qualification and a provider-driven agent lifecycle through the same public contract.}
\label{fig:controlled-forks}
\end{figure*}

\section{6. Agent interaction with
ChemWorld}\label{agent-interaction-with-chemworld}

The preceding experiments use deterministic execution to qualify the
worlds themselves. We finally test whether a qualified world can be
exposed to an autonomous agent through the same public contract while
preserving an evaluator-complete record. The goal is interface
integration, not a comparison of agent capability.

\subsection{6.1 Public agent interface and evaluator-complete
records}\label{public-agent-interface-and-evaluator-complete-records}

ChemWorld separates agent-facing information from evaluator-side
evidence. The agent receives the public task card, typed operations,
instruments, public observations, available resources and termination
and final-assay interfaces. It cannot directly inspect hidden material
properties, private constitutive-law parameters, private initialization
or complete simulator state. The evaluator retains the complete state
transitions, transaction outcomes, failure and resource consequences and
environment/action trajectory.

This information boundary permits evaluation without exposing the
mechanism being evaluated. ChemWorld groups 19 registered process
coordinates into terminal commitment, evidence acquisition,
evidence-conditioned action, resource deployment and outcome trajectory.
They remain separate coordinates rather than a composite score or a set
of uniformly higher-is-better metrics; their numerators, denominators,
null rules and boundary rules are specified in the process-coordinate
contract in Appendix D.1.

The resulting record can support subsequent comparisons of measurement
strategy, resource allocation, response to invalid actions and stopping
behavior. For example, an agent may choose between additional pH or
UV--visible evidence under a limited sample budget, or revise a typed
action after a structured rollback without losing committed history. The
deterministic cases in Section 5 qualify these interaction primitives,
but comparative policy behavior is outside the claims of this study.

\subsection{6.2 Independent-agent integration in a non-reference
world}\label{independent-agent-integration-in-a-non-reference-world}

World qualification and agent execution were kept as independent
experimental units so that agent success could not serve as evidence for
world correctness. The selected protocol-frozen non-reference world
combines reaction, thermal, distillation and observation components. A
deterministic 12-action path first qualified its construction, lifecycle
closure and exact replay. Only after world qualification was complete
did an independent agent enter the same public instrument contract and
complete a separate 15-action lifecycle (Fig. 3D).

The experimental interface supplied the public task card, typed tool
schemas, resource contract and explicit termination and final-assay
requirements. The agent received no private world fields and issued
every experimental operation, including termination and final assay. In
one uninterrupted experiment, it submitted 15 actions; all 15 committed
and closed the lifecycle, with no rollback, right-censoring or
undeclared private-field exposure.

The environment used 8,158.454 of 10,440 simulated process seconds, four
of four instrument uses and 0.00085 of 0.001 L sample. The
evaluator-complete record links each decision to its public observation,
hidden simulator consequence, transaction result and resource debit. The
entire 15-step submitted-action trace replayed with zero numerical
mismatch. An independent agent experiment can therefore enter the same
auditable execution and replay framework as the deterministic studies.
This demonstrates compatibility with the public instrument contract; it
is not an agent benchmark, a model comparison or a claim of general
chemical intelligence.

\section{7. Discussion and Conclusion}\label{discussion-and-conclusion}

\subsection{7.1 The chemical world as an experimental
variable}\label{the-chemical-world-as-an-experimental-variable}

ChemWorld's central methodological contribution is not a larger
collection of predefined tasks, but the ability to construct and control
the chemical world itself as an experimental variable. Component
composition and operating conditions specify what may vary; a unified
public experimental contract specifies what must remain comparable; and
transactional execution with evaluator-complete records preserves the
state, resource and failure consequences of every interaction.
Researchers can therefore alter process composition, continuous
operating conditions or hidden chemical and material laws while keeping
the agent-facing action and observation interface fixed, enabling
strictly matched experiments across different worlds.

The present study first establishes the reliability of this experimental
substrate. Tests of reference worlds, generated worlds, process modules,
component interfaces and invalid operations show that compatible
compositions retain common execution, resource and replay semantics
within the declared component and compatibility domain. Process-complete
lifecycles then show that the same contract supports distinct chemical
workflows and recovery after failure. Controlled forks further show that
one private law can be changed while the public experimental conditions,
action sequence and bound randomness remain fixed, allowing the
corresponding trajectory differences to be attributed to that
intervention. Finally, an independent agent completes a full lifecycle
in a non-reference world through the same public interface, without its
behavior being used to qualify the underlying world.

ChemWorld therefore provides a software experimental substrate for
controlled studies of world construction, world intervention and
agent--world interaction, rather than simply a larger fixed benchmark
suite.

\subsection{7.2 Implications for experimental-intelligence
research}\label{implications-for-experimental-intelligence-research}

Many agent benchmarks compress an experimental process into a success
rate, reward or terminal property. Such outcomes indicate whether a task
was completed, but reveal little about how the result was obtained.
ChemWorld instead retains the full process from evidence acquisition and
subsequent decisions through resource deployment, failure, recovery and
the terminal experimental outcome, making experimental strategy itself
available for study.

Under a limited resource budget, for example, researchers can compare
when an agent chooses to collect another measurement, when it stops, and
how it allocates sample and instrument capacity. After an invalid
operation, they can examine whether new observations alter the agent's
subsequent strategy. Within a controlled world fork, they can further
test whether an agent detects, adapts to or learns a changed hidden law.
The 19 process coordinates defined in this work provide a common record
for such analyses, but are intentionally not collapsed into a single
score, and this study does not use them to rank agents.

Controlled forks are especially important for these questions. Parent
and child worlds retain the same public task, operations, instruments,
resources, action sequence and bound randomness while differing in one
prespecified hidden law. Within the registered design and qualification
criteria, the associated state and observation differences can therefore
be attributed to that world intervention. Future work can build on this
substrate to study agent adaptation, law discovery and scientific
decision-making, but those claims require separately frozen agent
policies and evaluation protocols and lie beyond the present
qualification.

\subsection{7.3 Scope, extensibility and physical
validation}\label{scope-extensibility-and-physical-validation}

The conclusions of this study are limited to the declared component
vocabulary, compatibility rules and authored model domains.
Qualification establishes consistent executable semantics and internally
coherent behavior for these software models within their stated scope;
it does not establish that ChemWorld fully reproduces real chemical
systems. The current synthetic instruments are controlled software
observation models rather than digital twins calibrated to particular
devices. The coverage design systematically samples prespecified
parameter and composition spaces but does not exhaust chemical space or
all higher-order process interactions. Module fixtures and
directionality oracles primarily test consistency with the authored
models rather than an independent reference implementation or empirical
ground truth. Exact replay is likewise restricted to the bound software
identities, world identity and environment--action trace; it does not
imply policy re-execution, cross-platform numerical identity or
cross-version archival replay.

These boundaries do not require ChemWorld to remain fixed at its present
level of fidelity. Component interfaces separate experimental semantics
from individual model implementations, so experimentally calibrated
kinetic and thermodynamic laws, unit-operation models and
instrument-response models can in principle enter the same
world-construction, transaction and agent-interface framework. The
present study does not, however, establish that an independently
authored third-party module automatically inherits every system
guarantee without integration work. Each new model must be qualified
against the relevant interface and runtime semantics.

ChemWorld and physical self-driving laboratories are therefore
complementary experimental regimes rather than substitutes. Software
worlds support rapid repetition, strictly matched controls, targeted
intervention and complete process records; physical laboratories provide
real-material evidence and model calibration. A natural workflow is to
use software experiments to identify conditions, mechanisms and agent
behaviors that merit closer study, then return questions requiring real
chemical evidence to the physical laboratory.

In summary, ChemWorld establishes a programmable, repeatable and
auditable chemical experimental environment in which researchers can
systematically alter the world while preserving a common agent-facing
interface. This work qualifies that capability within the current
component and model domain and demonstrates complete experimental
lifecycles, single-law controlled counterfactuals and independent-agent
integration. Higher-fidelity chemical models, together with agent
adaptation, law learning and scientific discovery in changing worlds,
are natural directions enabled by this foundation.

\section{8. Data and Code
Availability}\label{data-and-code-availability}

A frozen, versioned public release associated with this study is openly
available under the MIT License in the
\href{https://github.com/sunyrain/ChemWorld-Public}{ChemWorld public
repository}; the frozen code-and-evidence snapshot referenced here is
tagged
\href{https://github.com/sunyrain/ChemWorld-Public/tree/v0.1.0}{v0.1.0}.
The release contains the executable code, publication configurations and
protocols, processed machine- and human-readable qualification evidence,
figure source files, replayable simulator environment--action
trajectories, the arXiv PDF and source bundle, and an offline manifest
verifier for the reported hashes and denominators. It is intentionally
separated from subsequent benchmark development. Credentials, raw
provider payloads, private seeds and local run caches are not included.

\clearpage
\bibliographystyle{unsrtnat}
\bibliography{references}

@article{boiko2023autonomous,
  author = {Boiko, Daniil A. and MacKnight, Robert and Kline, Ben and Gomes, Gabe},
  title = {Autonomous chemical research with large language models},
  journal = {Nature},
  volume = {624},
  pages = {570--578},
  year = {2023},
  doi = {10.1038/s41586-023-06792-0}
}

@article{bran2024augmenting,
  author = {Bran, Andres M. and Cox, Sam and Schilter, Oliver and Baldassari, Carlo and White, Andrew D. and Schwaller, Philippe},
  title = {Augmenting large language models with chemistry tools},
  journal = {Nature Machine Intelligence},
  volume = {6},
  pages = {525--535},
  year = {2024},
  doi = {10.1038/s42256-024-00832-8}
}

@article{szymanski2023alab,
  author = {Szymanski, Nathan J. and others},
  title = {An autonomous laboratory for the accelerated synthesis of inorganic materials},
  journal = {Nature},
  volume = {624},
  pages = {86--91},
  year = {2023},
  doi = {10.1038/s41586-023-06734-w}
}

@article{dai2024mobile,
  author = {Dai, Tianyi and Vijayakrishnan, Sriram and Szczypi{\'n}ski, Filip T. and others},
  title = {Autonomous mobile robots for exploratory synthetic chemistry},
  journal = {Nature},
  volume = {635},
  pages = {890--897},
  year = {2024},
  doi = {10.1038/s41586-024-08173-7}
}

@article{darvish2025organa,
  author = {Darvish, Kourosh and Skreta, Marta and Zhao, Yuchi and others},
  title = {{ORGANA}: A robotic assistant for automated chemistry experimentation and characterization},
  journal = {Matter},
  volume = {8},
  number = {2},
  year = {2025},
  doi = {10.1016/j.matt.2024.10.015}
}

@article{song2025chemagents,
  author = {Song, Tao and Luo, Man and Zhang, Xiaolong and others},
  title = {A Multiagent-Driven Robotic {AI} Chemist Enabling Autonomous Chemical Research On Demand},
  journal = {Journal of the American Chemical Society},
  volume = {147},
  number = {15},
  pages = {12534--12545},
  year = {2025},
  doi = {10.1021/jacs.4c17738}
}

@article{panapitiya2026autolabs,
  author = {Panapitiya, Gihan and Saldanha, Emily and Job, Heather and others},
  title = {{AutoLabs}: cognitive multi-agent systems with self-correction for autonomous chemical experimentation},
  journal = {Scientific Reports},
  volume = {16},
  pages = {19554},
  year = {2026},
  doi = {10.1038/s41598-026-45593-z}
}

@article{pilon2026robochemflex,
  author = {Pilon, Simone and Savino, Elia and Bayley, Oliver M. and others},
  title = {A flexible and affordable self-driving laboratory for automated reaction optimization},
  journal = {Nature Synthesis},
  year = {2026},
  doi = {10.1038/s44160-026-01053-0}
}

@article{vriza2026instruments,
  author = {Vriza, Aikaterini and Prince, Michael H. and Zhou, Tao and Chan, Henry and Cherukara, Mathew J.},
  title = {Operating advanced scientific instruments with {AI} agents that learn on the job},
  journal = {npj Computational Materials},
  volume = {12},
  pages = {160},
  year = {2026},
  doi = {10.1038/s41524-026-02005-0}
}

@article{felton2021summit,
  author = {Felton, Kobi C. and Rittig, Jan G. and Lapkin, Alexei A.},
  title = {Summit: Benchmarking Machine Learning Methods for Reaction Optimisation},
  journal = {Chemistry--Methods},
  volume = {1},
  pages = {116--122},
  year = {2021},
  doi = {10.1002/cmtd.202000051}
}

@article{hase2021olympus,
  author = {H{\"a}se, Florian and Aldeghi, Matteo and Hickman, Riley J. and Roch, Lo{\"i}c M. and Christensen, Melodie and Liles, Elena and Hein, Jason E. and Aspuru-Guzik, Al{\'a}n},
  title = {Olympus: a benchmarking framework for noisy optimization and experiment planning},
  journal = {Machine Learning: Science and Technology},
  volume = {2},
  pages = {035021},
  year = {2021},
  doi = {10.1088/2632-2153/abedc8}
}

@article{beeler2024chemgymrl,
  author = {Beeler, Chris and Subramanian, Sriram Ganapathi and Sprague, Kyle and others},
  title = {{ChemGymRL}: A customizable interactive framework for reinforcement learning for digital chemistry},
  journal = {Digital Discovery},
  volume = {3},
  pages = {742--758},
  year = {2024},
  doi = {10.1039/D3DD00183K}
}

@misc{bloor2024pcgym,
  author = {Bloor, Maximilian and Torraca, Jos{\'e} and Sandoval, Ilya Orson and others},
  title = {{PC-Gym}: Benchmark Environments For Process Control Problems},
  year = {2024},
  eprint = {2410.22093},
  archivePrefix = {arXiv},
  note = {Preprint, arXiv:2410.22093}
}

@inproceedings{malik2026made,
  author = {Malik, Shreshth A. and Doherty, Tiarnan and Tigas, Panagiotis and Razzak, Muhammed and Roberts, Stephen J. and Walsh, Aron and Gal, Yarin},
  title = {{MADE}: Benchmark Environments for Closed-Loop Materials Discovery},
  booktitle = {International Conference on Machine Learning},
  year = {2026},
  eprint = {2601.20996},
  archivePrefix = {arXiv}
}

@misc{xu2026scidisco,
  author = {Xu, Yucheng and Zhang, Keyi and Yu, Yuyang and Zhang, Min and Meng, Shiyuan and Chu, Pei and Tu, Zhongying},
  title = {Scaling Scientific Discovery Environments for Turn-Level Agentic {RL}},
  year = {2026},
  eprint = {2607.28990},
  archivePrefix = {arXiv},
  note = {Preprint, arXiv:2607.28990}
}

@inproceedings{jansen2024discoveryworld,
  author = {Jansen, Peter and C{\^o}t{\'e}, Marc-Alexandre and Khot, Tushar and Bransom, Erin and Mishra, Bhavana Dalvi and Majumder, Bodhisattwa Prasad and Tafjord, Oyvind and Clark, Peter},
  title = {{DiscoveryWorld}: A Virtual Environment for Developing and Evaluating Automated Scientific Discovery Agents},
  booktitle = {Advances in Neural Information Processing Systems},
  volume = {37},
  year = {2024},
  doi = {10.52202/079017-0324}
}

@misc{gandhi2025boxinggym,
  author = {Gandhi, Kanishk and Li, Michael Y. and Goodyear, Lyle and Bhatia, Agam and Li, Louise and Bhaskar, Aditi and Zaman, Mohammed and Goodman, Noah D.},
  title = {{BoxingGym}: Benchmarking Progress in Automated Experimental Design and Model Discovery},
  year = {2025},
  eprint = {2501.01540},
  archivePrefix = {arXiv},
  note = {Preprint, arXiv:2501.01540}
}

@misc{duan2025scigym,
  author = {Duan, Haonan and Lu, Stephen Zhewen and Harrigan, Caitlin Fiona and Desai, Nishkrit and Lu, Jiarui and Koziarski, Micha{\l} and Cotta, Leonardo and Maddison, Chris J.},
  title = {Measuring Scientific Capabilities of Language Models with a Systems Biology Dry Lab},
  year = {2025},
  eprint = {2507.02083},
  archivePrefix = {arXiv},
  note = {Preprint, arXiv:2507.02083}
}

@article{nagele2026sciexplorer,
  author = {N{\"a}gele, Maximilian and Marquardt, Florian},
  title = {Agentic Exploration of Physics Models},
  journal = {Physical Review X},
  volume = {16},
  pages = {031002},
  year = {2026},
  doi = {10.1103/xnqc-q6nt}
}

@inproceedings{zheng2026newtonbench,
  author = {Zheng, Tianshi and Tam, Kelvin Kiu-Wai and Nguyen, Newt Hue-Nam K. and others},
  title = {{NewtonBench}: Benchmarking Generalizable Scientific Law Discovery in LLM Agents},
  booktitle = {International Conference on Learning Representations},
  year = {2026},
  eprint = {2510.07172},
  archivePrefix = {arXiv}
}

@inproceedings{li2025labutopia,
  author = {Li, Rui and Hu, Zixuan and Qu, Wenxi and others},
  title = {{LabUtopia}: High-Fidelity Simulation and Hierarchical Benchmark for Scientific Embodied Agents},
  booktitle = {Advances in Neural Information Processing Systems, Datasets and Benchmarks Track},
  year = {2025},
  eprint = {2505.22634},
  archivePrefix = {arXiv}
}

@misc{wu2026labimus,
  author = {Wu, Yuhan and Jin, Zhao and Li, Tao and others},
  title = {{Labimus}: A Simulation and Benchmark for Humanoid Dexterous Manipulation in Chemical Laboratory},
  year = {2026},
  eprint = {2606.31037},
  archivePrefix = {arXiv},
  note = {Preprint}
}
\clearpage

\section{Appendix A. Qualification and Experimental
Protocols}\label{appendix-a.-qualification-and-experimental-protocols}

\subsection{A.1 Qualification and coverage
protocol}\label{a.1-qualification-and-coverage-protocol}

The authoritative protocol was frozen before execution. It specifies
component patterns, discrete axes and compatible pairs, continuous
bounds, seeds, ordered workflows and pass/fail criteria. Composition
qualification first checks dependencies, state ownership, unit
agreement, parameter domains, resource feasibility, exposed operations,
instrument availability and lifecycle closure. Invalid declarations must
return structured diagnostics without constructing an environment.

Discrete axes use pairwise covering rows, and continuous axes use seeded
Latin hypercube samples within authored bounds. Each pattern has one or
two ordered workflow targets. For every qualification unit, the report
records normalized construction input, compiler diagnostics, public
contract, submitted actions, transaction status, state-integrity checks,
events, resource preflight and outcome, public observations, termination
and exact replay. Missing receipts, non-finite quantities, unexpected
transaction outcomes, undeclared private-field exposure, denominator
drift or replay mismatch fail the unit.

Process-module qualification assigns reaction, thermal, phase,
separation, crystallization, distillation, continuous flow and
electrochemistry four probe classes each: zero-input bounded runtime,
valid low/high input, directionality and runtime invariants. Numerical
fixtures carry explicit tolerances; declared conceptual or synthetic
oracles are used where no independent numerical fixture is claimed.
Observation modules are checked separately for instrument availability,
sample accounting, bounded signals and the public/private boundary.
Invalid compiler probes cover missing dependencies, conflicting state
ownership, unit mismatch, invalid parameter domains, lifecycle holes and
resource impossibility.

\subsection{A.2 Process-time limits}\label{a.2-process-time-limits}

Process-time limits are derived by component pattern:

\[
t_{\max}=t_{\mathrm{required}}+t_{\mathrm{reserve}}+t_{\mathrm{repeat}}.
\]
\begin{table}[H]
\centering
\small
\caption{\textbf{Protocol-frozen process-time limits.}}
\label{tab:process-time-limits}
\begin{tabularx}{\columnwidth}{@{}Yr@{}}
\toprule
Pattern & Limit (s) \\
\midrule
Phase observation & 0 \\
Reaction--thermal observation & 3,600 \\
Phase separation & 1,860 \\
Reaction crystallization & 11,100 \\
Reaction distillation & 10,440 \\
Continuous flow & 7,200 \\
Electrochemistry & 5,400 \\
Reaction--phase separation & 7,500 \\
\bottomrule
\end{tabularx}
\end{table}

The required term sums the upper bounds of necessary timed stages, the
reserve covers authored quench and transfer operations, and the repeat
term follows explicit per-operation repeat limits. Preflight rejects an
action if either cumulative process time or a repeat limit would be
exceeded.

\subsection{A.3 Deterministic use-case
protocol}\label{a.3-deterministic-use-case-protocol}

The cases, seeds, submitted-action sequences and terminal requirements
in Table \ref{tab:deterministic-protocol} were fixed before execution.
Every case requires explicit termination and final assay. The
failure--recovery case additionally fixes a runtime-precondition failure
at its first submitted action and requires continuation from the
restored committed state.

\begin{table}[H]
\centering
\fontsize{6.7}{7.5}\selectfont
\renewcommand{\arraystretch}{0.88}
\caption{\textbf{Protocol-frozen deterministic use cases.} Submitted actions include termination and final assay.}
\label{tab:deterministic-protocol}
\begin{tabularx}{\columnwidth}{@{}L{0.20\columnwidth}YL{0.12\columnwidth}L{0.17\columnwidth}@{}}
\toprule
Case & Public identity & Seed / actions & Planned failure \\
\midrule
U01 & reaction-to-crystallization & 0 / 12 & none \\
U02 & composed-equilibrium-characterization-demo & 0 / 5 & none \\
U03/E01 & composed-reaction-purification-demo & 0 / 19 & step-1 runtime rollback \\
U06-flow & flow-reaction-optimization & 0 / 8 & none \\
U06-electro & electrochemical-conversion & 0 / 11 & none \\
U06-distillation & reaction-to-distillation & 0 / 12 & none \\
U06-partition & partition-discovery & 0 / 10 & none \\
U06-crystallization & reaction-to-crystallization & 1 / 12 & none \\
\bottomrule
\end{tabularx}
\end{table}

\subsection{A.4 Controlled-fork acceptance
criteria}\label{a.4-controlled-fork-acceptance-criteria}

For an aligned checkpoint value \(p\) in the parent and \(c\) in the
child, the acceptance oracle records

\[
\delta=c-p,\qquad \Delta=|\delta|,\qquad
r=\frac{\Delta}{\max(|p|,|c|,s_0)},
\]

where \(s_0\) is the registered positive scale floor. A fork passes only
if the registered direction, absolute threshold, relative threshold,
public-contract invariance, single-target lineage, same-sequence
executability and exact-replay criteria are all satisfied.

\subsection{A.5 Public-boundary and exact-replay
checks}\label{a.5-public-boundary-and-exact-replay-checks}

Public-boundary checks scan agent-facing task cards, observations and
histories for undeclared evaluator-owned fields, private identifiers and
absolute private paths. Inferential information obtained through
task-declared measurements is not counted as direct exposure.

Replay reconstructs the bound world from its normalized contract,
runtime, mechanism, observation and scoring identities together with
recorded seeds and interventions, then resubmits every typed action in
recorded order. Field-level comparison covers rewards, public
observations, termination and truncation flags, operation types,
transaction status, rollback reason, affected resource-ledger fields,
world events, state-delta summaries and state-integrity checks. The
replay result is reconciled with resource ledgers and rollback receipts;
exact replay requires zero numerical tolerance and no field mismatch.

\clearpage
\nobalance
\twocolumn[
\begin{@twocolumnfalse}
\section{Appendix B. World Registry and Component Library}\label{appendix-b}
\subsection{B.1 Reference-task registry}\label{appendix-b.1-reference-task-registry}
\centering
\scriptsize
\captionsetup{hypcap=false}
\captionof{table}{\textbf{Complete reference-task registry.} Seed counts define the scenario identities exercised for each task.}
\label{tab:reference-registry}
\begin{tabularx}{\textwidth}{@{}L{0.22\textwidth}L{0.32\textwidth}rY@{}}
\toprule
Reference task identity & Component topology & World seeds & Public operation / instrument summary \\
\midrule
electrochemical-conversion & reaction + electrochemistry + observation & 5 & 6 operations; pH, UV--visible, final assay \\
equilibrium-characterization & reaction + thermal + observation & 5 & 9 operations; pH, UV--visible, final assay \\
flow-reaction-optimization & reaction + continuous flow + observation & 5 & 7 operations; HPLC, GC, UV--visible, final assay \\
low-budget-characterization & reaction + thermal + observation & 3 & 9 operations; HPLC, GC, UV--visible, final assay \\
partition-discovery & reaction + phase + separation + observation & 5 & 9 operations; HPLC, GC, UV--visible, final assay \\
public-private-generalization & reaction + thermal + observation & 5 & 9 operations; HPLC, GC, UV--visible, final assay \\
purity-yield-tradeoff & reaction + thermal + phase + separation + observation & 5 & 18 operations; HPLC, GC, UV--visible, final assay \\
reaction-mechanism-explanation & reaction + thermal + observation & 3 & 9 operations; HPLC, GC, UV--visible, final assay \\
reaction-optimization-standard & reaction + thermal + observation & 5 & 9 operations; HPLC, GC, UV--visible, final assay \\
reaction-safety-constrained & reaction + thermal + observation & 5 & 9 operations; HPLC, GC, UV--visible, final assay \\
reaction-to-assay & reaction + thermal + observation & 1 & 9 operations; HPLC, GC, UV--visible, final assay \\
reaction-to-crystallization & reaction + thermal + crystallization + observation & 5 & 12 operations; HPLC, final assay \\
reaction-to-distillation & reaction + thermal + distillation + observation & 5 & 12 operations; HPLC, GC, UV--visible, final assay \\
reaction-to-purification & reaction + thermal + phase + separation + observation & 5 & 18 operations; HPLC, GC, UV--visible, final assay \\
tool-agent-planning & reaction + thermal + phase + separation + observation & 2 & 18 operations; HPLC, GC, UV--visible, final assay \\
\bottomrule
\end{tabularx}
\par\medskip
\subsection{B.2 Component-pattern library}\label{appendix-b.2-component-pattern-library}
\centering
\small
\captionof{table}{\textbf{Representative component-pattern library.} Reusable combinations expose a common operation and instrument surface across single-process and multistage worlds.}
\label{tab:component-pattern-library}
\begin{tabularx}{\textwidth}{@{}L{0.19\textwidth}L{0.20\textwidth}YL{0.19\textwidth}@{}}
\toprule
Component pattern & Principal state or process & Representative public operations & Representative instruments \\
\midrule
Phase + observation & bounded phase/equilibrium state & add solvent, add reagent, measure, terminate & pH, UV--visible, final assay \\
Reaction + thermal + observation & batch reaction and temperature history & add, heat, quench, sample & HPLC, GC, final assay \\
Phase + separation + observation & phase formation and transfer & mix, settle, separate, wash, transfer & HPLC, final assay \\
Reaction + thermal + crystallization + observation & reaction followed by solid formation & heat, seed, cool, filter & HPLC, particle sizing, final assay \\
Reaction + thermal + distillation + observation & reaction, evaporation and fractionation & heat, quench, evaporate, distil, collect & HPLC, GC, final assay \\
Reaction + thermal + continuous flow + observation & flow, residence time and conversion & set flow, set temperature, run, sample & HPLC, GC, final assay \\
Reaction + electrochemistry + observation & potential/current-driven conversion & set potential/current, electrolyse, sample & voltammetry, HPLC, final assay \\
Reaction + thermal + phase + separation + observation & multistage reaction and purification & react, quench, separate, wash, concentrate, transfer & HPLC, GC, final assay \\
\bottomrule
\end{tabularx}
\end{@twocolumnfalse}
]

\clearpage
\twocolumn[
\begin{@twocolumnfalse}
\subsection{B.3 Component model cards and extension points}\label{appendix-b.3-component-model-cards}
\centering
\fontsize{7.6}{8.4}\selectfont
\renewcommand{\arraystretch}{0.88}
\captionsetup{hypcap=false}
\captionof{table}{\textbf{Component model cards and extension points.} Each module declares its runtime formulation, authored model domain, qualification oracle and extension interface for future alternative or empirically calibrated implementations.}
\label{tab:component-model-cards}
\begin{tabularx}{\textwidth}{@{}L{0.125\textwidth}L{0.185\textwidth}L{0.18\textwidth}L{0.20\textwidth}Y@{}}
\toprule
Component & Runtime formulation & Principal authored domain & Qualification oracle & Intended use and extension path \\
\midrule
Reaction & stoichiometric mass-action network with Arrhenius temperature dependence & authored reaction families and bounded batch temperature/time & exact amount fixtures, monotonic response, material closure and runtime invariants & reaction-law extension point for future named-reaction calibration \\
Thermal & dynamic batch heat-release and jacket-energy balance & bounded temperature, duration, vessel pressure and volume & temperature/energy finiteness, bounds, event propagation and ledger reconciliation & thermal-contract extension point for future equipment-specific coefficients \\
Phase & stability-gated, activity-corrected liquid--liquid equilibrium with TPD-style diagnostics & declared phase identities, volumes and composition ranges & phase/material balance, directional partition response and state identity & activity-model extension point for future compound-specific thermodynamics \\
Separation & settling, entrainment, wash and transfer coupled to the phase model & bounded mix/settle time, extractant/wash volume and transfer fraction & amount/unit conservation, transfer identity and expected directional response & separation-interface extension point for future hardware-scale transport \\
Crystallization & van't Hoff solubility with seed, nucleation/growth cohorts, impurity occlusion and CSD summaries & bounded seed mass, cooling temperature and cooling time & material closure, solubility-direction checks, CSD and runtime-invariant receipts & law extension point for future calibrated nucleation and growth models \\
Distillation & bubble-gated, duty-limited VLE/Fenske fractionation with material and energy ledgers & bounded temperature/time, reflux ratio, fraction count and collected fraction & mass/energy closure, fraction identity, recovery/purity directions and equipment limits & operation-contract extension point for future compound and column models \\
Continuous flow & geometry-resolved plug-flow reactor with residence time, distributed thermal boundary and pressure drop & bounded flow, residence time and temperature & conversion direction, mass closure, pressure/geometry and solver diagnostics & runtime extension point for future reactor-specific transport or control models \\
Electrochemistry & Nernst potential, Butler--Volmer kinetics, limiting current, Randles transient and Faraday accounting & bounded potential, current and electrolysis time & charge/material closure, signed work, selectivity and limiting-current checks & electrochemical-law extension point for future material- and cell-specific parameters \\
Observation & state-coupled synthetic pH, UV--visible, HPLC, GC and final-assay contracts & task-declared instruments, sample and use budgets & instrument availability, sample consumption, finite/bounded signal and non-omniscience checks & instrument-schema extension point for future empirical response models \\
\bottomrule
\end{tabularx}
\par\medskip
\section{Appendix C. Coverage and Qualification Results}\label{appendix-c}
\noindent This appendix reports the complete frozen construction-coverage and qualification census referenced in Section 4.
\subsection{C.1 Topology and identity decomposition}\label{appendix-c.1-topology-and-identity}
\centering
\small
\captionsetup{hypcap=false}
\captionof{table}{\textbf{Topology and identity decomposition of the generated block.} Topology novelty compares exact component sets with the 15-task reference registry; task--world novelty uses exact registered identities.}
\label{tab:novelty-decomposition}
\begin{tabularx}{\textwidth}{@{}L{0.33\textwidth}rL{0.22\textwidth}Y@{}}
\toprule
Generated group & Cases & Topology relation & Exact task--world relation \\
\midrule
phase--observation & 6 & absent from reference registry & zero overlap implied by component-set difference \\
phase--separation--observation & 6 & absent from reference registry & zero overlap implied by component-set difference \\
reaction--thermal--continuous-flow--observation & 6 & absent from reference registry & zero overlap implied by component-set difference \\
reaction--thermal--distillation--observation & 8 & registered topology & zero exact overlap in the protocol-frozen non-reference block \\
four remaining generated patterns & 26 & registered topologies & used for reference-topology coverage; no separate non-reference identity claim \\
\bottomrule
\end{tabularx}
\end{@twocolumnfalse}
]

\clearpage
\twocolumn[
\begin{@twocolumnfalse}
\subsection{C.2 Frozen coverage design}\label{appendix-c.2-frozen-coverage-design}
\centering
\scriptsize
\captionsetup{hypcap=false}
\captionof{table}{\textbf{Protocol-frozen coverage design.} Bounds are inclusive authored domains; ``none'' denotes a purely discrete design.}
\label{tab:coverage-design}
\begin{tabularx}{\textwidth}{@{}L{0.21\textwidth}rYL{0.10\textwidth}r@{}}
\toprule
Pattern & Seed & Continuous axes and bounds & Workflows & Cases \\
\midrule
phase--observation & 101 & none & 1 & 6 \\
reaction--thermal--observation & 102 & heat 350--390 K; duration 600--1,800 s & 2 & 6 \\
phase--separation--observation & 103 & phase 0.010--0.020 L; extractant 0.010--0.025 L; mix 60--300 s; settle 120--600 s & 2 & 6 \\
reaction--thermal--crystallization--observation & 104 & reaction 350--390 K, 600--1,800 s; seed 0.002--0.010 g; cooling 275--305 K, 900--3,600 s & 2 & 6 \\
reaction--thermal--distillation--observation & 105 & reaction 350--390 K, 600--1,800 s; evaporation 325--345 K, 300--900 s; distillation 350--390 K, 900--2,400 s; reflux 1.0--3.0; transfer 0.65--0.95 & 2 & 8 \\
reaction--thermal--continuous-flow--observation & 106 & flow 0.5--5.0 mL min$^{-1}$; residence 60--600 s; temperature 330--390 K & 2 & 6 \\
reaction--electrochemistry--observation & 107 & potential 0.5--1.8 V; current 25--150 mA; electrolysis 300--1,800 s & 2 & 7 \\
reaction--thermal--phase--separation--observation & 108 & reaction 350--390 K, 600--1,800 s; phase/extractant 0.010--0.020/0.010--0.025 L; mix 60--300 s; settle 120--600 s; wash 0.003--0.010 L; concentrate 300--900 s; transfer 0.65--0.95 & 2 & 7 \\
\bottomrule
\end{tabularx}
\par\medskip
\subsection{C.3 Qualification census}\label{appendix-c.3-qualification-census}
\centering
\small
\captionof{table}{\textbf{Qualification census.} Every registered execution, probe, deterministic-use and controlled-fork denominator completed without missing receipts or unexpected outcomes.}
\label{tab:qualification-census}
\begin{tabularx}{\textwidth}{@{}YrrY@{}}
\toprule
Qualification unit & Passed & Denominator & Failure classes \\
\midrule
Reference task--world units & 64 & 64 & 0 \\
Complete reference recipes & 1,786 & 1,786 & 0 \\
Coverage-generated compositions & 52 & 52 & 0 \\
Protocol-frozen non-reference reaction--distillation compositions & 8 & 8 & 0 \\
Invalid action probes & 192 & 192 & 0 unexpected outcomes \\
Module probes & 32 & 32 & 0 \\
Cross-module interface paths & 7 & 7 & 0 \\
Invalid compile mutants & 7 & 7 & 0 unexpected constructions \\
Deterministic use cases & 8 & 8 & 0 \\
Deterministic submitted actions & 89 & 89 & 0 missing receipts \\
Controlled fork pairs & 6 & 6 & 0 \\
Controlled fork traces & 24 & 24 & 0 \\
\bottomrule
\end{tabularx}
\end{@twocolumnfalse}
]

\clearpage
\twocolumn[
\begin{@twocolumnfalse}
\section{Appendix D. Agent-Facing Evaluation Records}\label{appendix-d}
\subsection{D.1 Process-coordinate dictionary}\label{appendix-d.1-process-coordinate-dictionary}
\noindent The 19 coordinates are campaign-level descriptions, not a composite score.
\(N_p\), \(N_c\), \(N_a\), \(N_d\) and \(N_m\) denote planned, closed, assayed,
discarded and measured lifecycles, with \(N_c=N_a+N_d\) because the frozen lifecycle
contract closes only by assay or discard. Undefined conditional quantities remain null;
they are never replaced by zero.
\par\medskip
\centering
\fontsize{7.6}{8.2}\selectfont
\renewcommand{\arraystretch}{0.84}
\captionsetup{hypcap=false}
\captionof{table}{\textbf{The 19 process coordinates.} Numerators, denominators, null rules and interpretation are fixed by the process-profile contract. None is interpreted as a globally better direction unless explicitly stated.}
\label{tab:process-coordinates}
\begin{tabularx}{\textwidth}{@{}L{0.16\textwidth}L{0.24\textwidth}L{0.20\textwidth}Y@{}}
\toprule
Coordinate & Definition & Null or boundary rule & Interpretation \\
\midrule
\multicolumn{4}{@{}l}{\textit{Terminal commitment}} \\
Closed lifecycle fraction & $N_c/N_p$ & defined as 0 when $N_p>0$ and none closes & completion gate; no preferred assay/discard mix \\
Assay commitment fraction & $N_a/N_c$ & null when $N_c=0$ & fraction of closed lifecycles committed to final assay \\
Discard fraction & $N_d/N_c$ & null when $N_c=0$ & fraction of closed lifecycles deliberately discarded \\
\addlinespace
\multicolumn{4}{@{}l}{\textit{Evidence acquisition}} \\
Measured lifecycle fraction & $N_m/N_c$ & null when $N_c=0$ & prevalence of at least one committed non-final measurement \\
Instrument uses per closed lifecycle & committed non-final measurements $/N_c$ & null when $N_c=0$ & measurement intensity \\
First-measurement timing & mean of preceding attempts $/$ attempts through termination & null when $N_m=0$ & lower values mean earlier evidence acquisition, not higher quality \\
\addlinespace
\multicolumn{4}{@{}l}{\textit{Evidence-conditioned action}} \\
Post-measure continuation prevalence & lifecycles with a later committed physical operation $/N_c$ & null when $N_c=0$ & prevalence of evidence followed by further investment among closed lifecycles \\
Post-measure operations per closed lifecycle & later committed physical operations $/N_c$ & null when $N_c=0$ & deployment intensity after first evidence \\
Threshold-eligible fraction & lifecycles with the frozen diagnostic and finite signal $/N_c$ & null when $N_c=0$ & denominator gate for the diagnostic decision rule \\
Evidence-to-terminal concordance & terminal choices matching the frozen signal rule $/$ eligible lifecycles & null when no lifecycle is eligible & agreement between declared evidence rule and assay/discard decision \\
\addlinespace
\multicolumn{4}{@{}l}{\textit{Resource deployment}} \\
Attempted operations per closed lifecycle & charged operation attempts $/N_c$ & null when $N_c=0$ & includes validation failures and transactional rollbacks \\
Committed operations per closed lifecycle & committed typed operations $/N_c$ & null when $N_c=0$ & installed operation intensity \\
Cost per closed lifecycle & campaign cost-ledger debit $/N_c$ & null when $N_c=0$ & includes declared failed-attempt charges \\
Risk debit per closed lifecycle & campaign risk-ledger debit $/N_c$ & null when $N_c=0$ & resource-card-specific risk deployment \\
\addlinespace
\multicolumn{4}{@{}l}{\textit{Outcome trajectory}} \\
Global-best discovery fraction & $(j^\star-1)/(N_a-1)$, $j^\star\in\{1,\ldots,N_a\}$ & null for no assay; defined as 0 for one assay & lower values mean earlier discovery of the observed best \\
Online incumbent retention & later assays retaining at least 90\% of the prior incumbent $/(N_a-1)$ & null when $N_a<2$ & stability after the first assay \\
Maximum incumbent drawdown & $\max(\text{prior incumbent}-\text{next assay},0)$ & null when $N_a<2$ & largest observed loss from the running best \\
Loss-episode recovery rate & recovered loss episodes $/$ observed loss episodes & null when no loss episode occurs; terminal unresolved losses count unrecovered & recovery after an observed loss \\
Terminal-to-best retention & terminal assayed score $/$ observed best assayed score & null when no positive assay exists & closeness of the terminal assay to the observed best \\
\bottomrule
\end{tabularx}
\par\medskip
\raggedright
\normalsize
\subsection{D.2 Independent-agent protocol}\label{appendix-d.2-independent-agent-protocol}
\noindent The agent experiment consists of one complete lifecycle on the protocol-frozen
non-reference reaction--distillation world. The agent interacts exclusively through the
public instrument interface, is limited to at most 16 submitted actions, and must explicitly
issue termination and final assay. The run passes only if the lifecycle closes correctly,
all submitted actions commit, action and trajectory records agree, resource constraints are
respected, no private fields are exposed, and the complete submitted-action trace replays
exactly.
\end{@twocolumnfalse}
]

\balance

\end{document}